\documentclass[letterpaper]{article} 
\usepackage{aaai25}
\usepackage{times}  
\usepackage{helvet}  
\usepackage{courier}  
\usepackage[hyphens]{url}  
\usepackage{graphicx} 
\usepackage{natbib}  
\usepackage{caption} 
\usepackage{algorithm}
\usepackage{algorithmic}

\usepackage{here}
\usepackage{xcolor}

\usepackage{newfloat}
\usepackage{listings}
\DeclareCaptionStyle{ruled}{labelfont=normalfont,labelsep=colon,strut=off} 
\floatstyle{ruled}
\newfloat{listing}{tb}{lst}{}
\floatname{listing}{Listing}
\title{Brain Hematoma Marker Recognition Using Multitask Learning: SwinTransformer and Swin-Unet}
\author{
    Kodai Hirata,
    Tsuyoshi Okita
}
\affiliations{
    \textsuperscript{\rm 1}Kyushu Institute of Technology, Department of Artificial Intelligence\\ 

    680-4 Kawazu, Iizuka-shi, Fukuoka, 820-8502, JAPAN.
}

\usepackage{bibentry}

\begin{document}

\maketitle

\begin{abstract}
This paper proposes a method MTL-Swin-Unet which is multi-task learning using transformers for classification and semantic segmentation. For sprious-correlation problems,
this method allows us to enhance the image representation with
two other image representations: representation obtained by
semantic segmentation and representation obtained by image reconstruction.
In our experiments, the proposed method outperformed in F-value measure than other classifiers when the test data included slices from the same patient (no covariance shift). Similarly, when the test data did not include slices
from the same patient (covariance shift setting), the proposed method
outperformed in AUC measure.
\end{abstract}

%

\section{Introduction}
Image recognition has transitioned from convolutional neural network-based approaches to vision transformer-based methods, significantly improving performance through the use of pretraining. In the case of images, pretraining is performed on objects from datasets such as ImageNet \cite{deng2009imagenet} and CoCo. Models pretrained in this manner are then utilized for transfer learning and fine-tuning. By adopting this approach, when the pretrained image data domain differs from the target domain, domain adaptation techniques—such as narrowing the gap between domains—are employed. While this approach can yield the desired results when the target domain is similar to the pretraining objects, it may not address cases where the target domain data exhibits spurious correlation \cite{sagawa2020investigation}.

The phenomenon known as ‘spurious correlation’{\cite{sagawa2020investigation} occurs when a machine learning model inadvertently constructs associations between objects that are not the intended targets. Even though the purpose is to recognize a specific object, other unrelated objects become mistakenly associated with the target. For instance, consider images of waterfowl: if the focus is on waterfowl, the background often includes water bodies such as lakes, ponds, rivers, or seas—simply because waterfowl are frequently found in such environments. In unfortunate cases, the model may mistakenly learn to associate the water in the background with the waterfowl, rather than correctly identifying the intended subject. This phenomenon is referred to as spurious correlation. In the context of image recognition, there are methods to verify prediction rationale. By examining color-coded images generated using Grad-CAM \cite{DBLP:journals/corr/SelvarajuDVCPB16} or attention weights, one can assess whether the model has fallen into spurious correlations. If the trained model has learned waterfowl, those birds will be highlighted; if it has learned water-related objects, a different area will be emphasized.

The focus of this paper is to classify a specific type of lesion known as ‘hypodensity’ in brain images. In addition to this target hypodensity marker, there are three other lesion markers in the setting. Specifically, when classifying the positive class (hypodensity marker), the negative class includes images of three different types of lesions, as well as numerous non-lesion images. Due to this configuration, there is a tendency for pseudo-correlation between positive class images containing lesions and negative class images without lesions. When we experimented with constructing a classifier, it became evident that areas such as the skull and central brain, apart from the lesions, were falsely colored, indicating pseudo-correlation. Our classification problem in this paper corresponds to this highly specialized scenario—not solely because it involves medical images, but rather because the target region is specifically the lesion itself. Please be aware of this distinction.

Approaches to address the spurious correlation problem have included methods based on weighting \cite{kirichenko2023layer}, sampling \cite{idrissi2022simple}, data augmentation \cite{hwang2022selecmix}, post-calibration \cite{kirichenko2023layer}, and worst-group error \cite{sagawa2020investigation}. 

In this paper,
our aim is to adopt the multitask learning to enhance the overall accuracy, by refining image representations using different objectives. In multi-task learning setting, the network is optimized by combining the losses obtained from multiple of tasks. The representations obtained from multiple tasks often lead to the improved generalization performance.

In order to do so, our first interest was the segmentation task. We luckily have at hand the masked images as well for brain hematoma segmentation.
We deployed this for jointly learning image recognition and segmentation.
By visualizing the areas the model focuses on with Grad-CAM, we found that the model strongly focuses on the hematomas.
We noticed that we suffer from the spurious correlation task.

\section{Related Works}
\subsection{Multi-task Learning with Transformer}
As mentioned in the previous chapter, the resolution-changing property of the Swin Transformer is well-suited for segmentation tasks.
Therefore, Swin Transformer is also used in multi-task learning, including segmentation tasks. MulT \cite{bhattacharjee2022mult} predicts multiple pixel-level tasks such as segmentation tasks, depth estimation, reshading, and edge detection using the U-shaped structure of Swin Transformer. In the research on multi-task Swin UNETR \cite{grzeszczyk2022multi}, Swin Transformer is used as an encoder for multi-task learning of medical image segmentation tasks and classification tasks.

\subsection{Joint Learning}
In general multi-task learning, shared parameters are updated by learning multiple tasks simultaneously. However, there is also research on learning methods that sequentially use representations obtained from one task for another task. In the study \cite{belharbi2021deep}, the model is first trained for the classification task. Then, the parameters of the classification phase are fixed, and the output of the backbone is used to train the segmentation task.

\section{Method}
In this chapter, we describe how we constructed the model to combine multiple tasks. We conducted learning using two methods: multi-task learning and joint learning. In multi-task learning, we extended Swin-Unet, which is designed for semantic segmentation, by adding classification and reconstruction tasks. For joint learning, we performed semantic segmentation using Swin-Unet and then utilized the encoder representations from SwinTransformer for the classification task. In our experiments, we refer to the first method as MTL-Swin-Unet and the second method as Joint-SwinTransformer.

\subsection{Multi-task Learning Using Swin-Unet}
\subsubsection{Architecture}
The architecture of the first proposed method is shown in Figure \ref{fig:mtl-architecture}. We extend the Swin-Unet, designed for semantic segmentation, by adding reconstruction and classification tasks. Each task shares the Swin Transformer encoder. The encoder performs gradual downsampling of the image, generating hierarchical representations. For the segmentation and reconstruction tasks, we utilize a Swin Transformer decoder with a symmetric arrangement to the encoder. The decoder reconstructs the input image resolution using representations extracted by the encoder and representations from each hierarchical level sent via skip connections. In the classification task, we perform classification using a linear layer with the representations obtained from the encoder.

\begin{figure*}[t]
\begin{center}
\includegraphics[width=15cm]{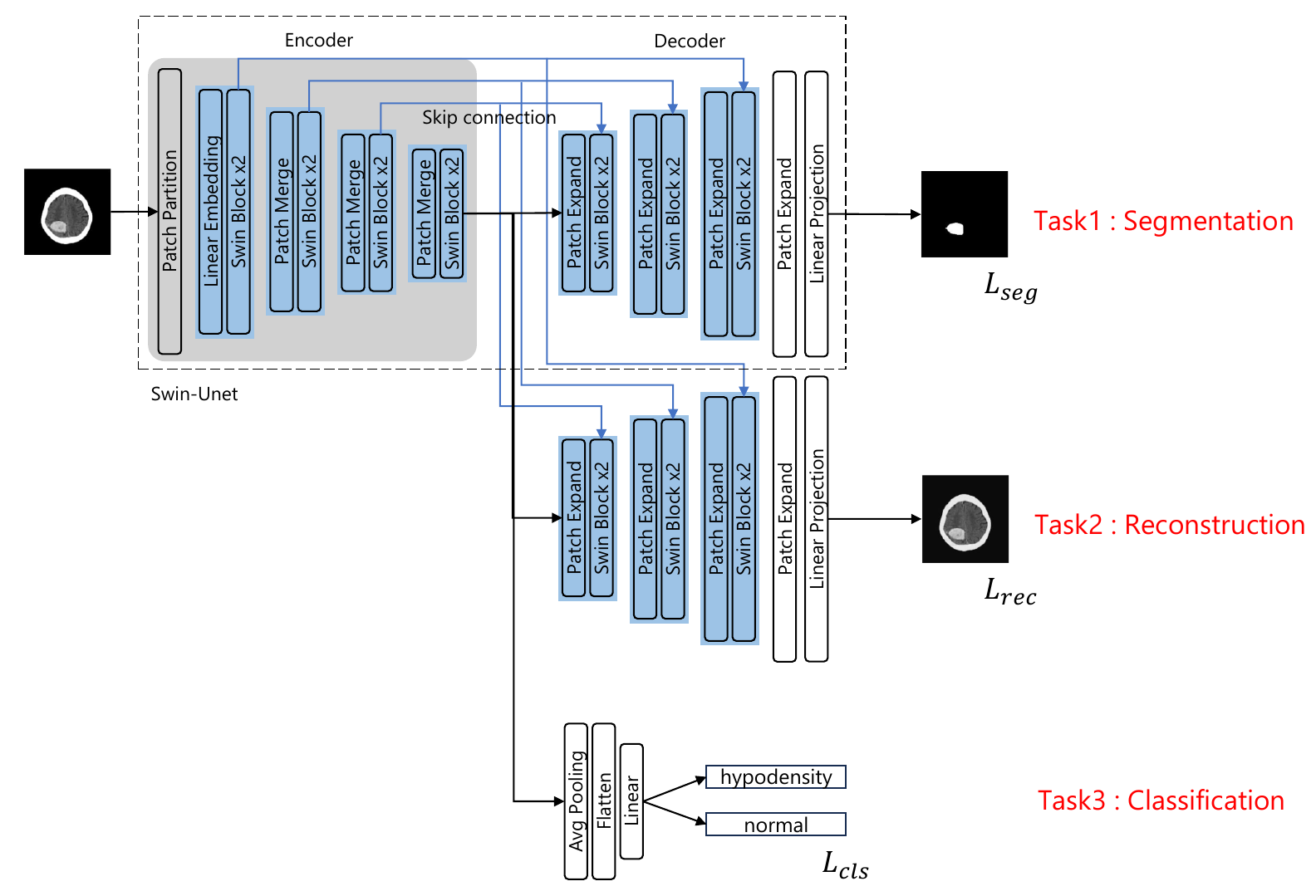}   
\caption{Architecture of MTL-Swin-Unet}
\label{fig:mtl-architecture}
\end{center}
\end{figure*}

\paragraph{Decoder for Segmentation and Image Reconstruction Tasks}
The decoders for both the segmentation task and image reconstruction task utilize the decoder employed in Swin-Unet for semantic segmentation. These decoders exhibit a symmetric structure with the Swin Transformer encoder. 

At the beginning of each stage, the representations obtained by the encoder undergo upsampling using patch expansion layers. In this layer, the channel dimension of the input feature is first expanded by a linear layer to double its size \((h,w,C)\to(h,w,2C)\). Next, the rearrange process performs shape transformation, enabling upsampling from the channel dimension to the resolution dimension. \((h,w,2C)\to(2h,2w,\frac{C}{2})\). Through this process, we symmetrically reduce the dimensionality while increasing the resolution, in contrast to the patch merge layer.

Before the SwinTransformer block, skip connections are applied similarly to Unet. The representations from the same layer in the encoder are preserved and combined along the channel dimension. Finally, the dimensions that were doubled by a linear layer are restored to their original dimensions. This process helps compensate for spatial information lost during downsampling.

\paragraph{Classification Head} \label{sec:classification head}
In classification tasks, global average pooling is applied to the final output of the encoder, followed by a linear layer. Models such as ViT \cite{dosovitskiy2020image} and DeiT \cite{touvron2021training} use the final output as a CLS-token. In this context, where patches serve as the output, global average pooling is used to aggregate representations.

\subsubsection{Loss Function}
Semantic segmentation tasks predict objects within an image at the pixel level. Due to the complexity of segmentation tasks, a weighted sum of two loss functions is commonly used. The first loss function is the cross-entropy function, which directly optimizes the segmentation task’s loss. It is expressed by the following equation.
\begin{equation}
L_{ce}(p,q) = -\sum_{i} p_{i}\log q_{i}
\end{equation}
Here, $p$ represents the ground truth label, and $q$ represents the inference result. However, when the objects in the image are small, it can lead to imbalanced data, making optimization challenging. Therefore, as a second loss function, Dice loss is employed. Dice loss calculates the degree of overlap between the ground truth region and the inferred region, as expressed by the following equation.

\begin{equation}
L_{dice}(p,q) = 1 - \frac{2\sum_{i}p_{i}q_{i}}{\sum_{i}(p_{i}+q_{i})}
\end{equation}
The final loss for semantic segmentation tasks is a weighted sum of two loss functions, expressed by the following equation.
\begin{equation}
L_{seg} = \lambda_{ce}\cdot L_{ce} + \lambda_{dice} \cdot L_{dice}
\end{equation}
Here, \(\lambda\)represents the weight for each task, and in the experiments, we use \(\lambda_{ce}=0.4\) and \(\lambda_{dice}=0.6\). The image reconstruction task trains the decoder’s output to predict the input image. For the reconstruction task, mean squared error (MSE) is employed. MSE is expressed by the following equation.

\begin{equation}
L_{rec}(p,q) = \frac{1}{n}\sum_{i}(p_{i}-q_{i})^2
\end{equation}
In classification tasks, we calculated the loss \(L_{cls}\) using the cross-entropy function, similar to the first loss function in segmentation tasks. Multi-task learning typically optimizes by taking a weighted sum of the losses for each task to find the minimum value. The final loss is computed as a weighted sum of the individual task losses, as expressed by the following equation.

\begin{equation}
L = \lambda_{cls}\cdot L_{cls} + \lambda_{seg} \cdot L_{seg} + \lambda_{rec} \cdot L_{rec}
\end{equation}
In the experiments, for a multi-task learning scenario with three tasks, we use weights \(\lambda_{cls}=0.3\), \(\lambda_{seg}=0.4\), and \(\lambda_{rec}=0.4\). Additionally, for a two-task multi-task learning scenario, we use either \(\lambda_{cls}=0.4\), \(\lambda_{seg}=0.6\), or \(\lambda_{cls}=0.4\), \(\lambda_{rec}=0.6\).
Finally, when performing both segmentation and classification tasks simultaneously, there are cases where labels exist for classification but not for segmentation. For instance, in images without hemorrhages, there are no segmentation mask images available. Therefore, the segmentation loss is computed using only images with existing segmentation masks, averaged over batches.

\subsection{Joint Learning Using Swin-Unet}
\subsubsection{Architecture}
The architecture of the second proposed method is shown in Figure \ref{fig:joint-architecture}. This learning approach is based on the joint learning study. First, we perform the segmentation task using Swin-Unet to predict hemorrhage pixels. Next, we freeze the parameters of the encoder trained with Swin-Unet. Finally, we combine the representations from the regular SwinTransformer encoder and the frozen encoder to utilize them for the classification task.


\begin{figure*}[t]
\begin{center}
\includegraphics[width=13cm]{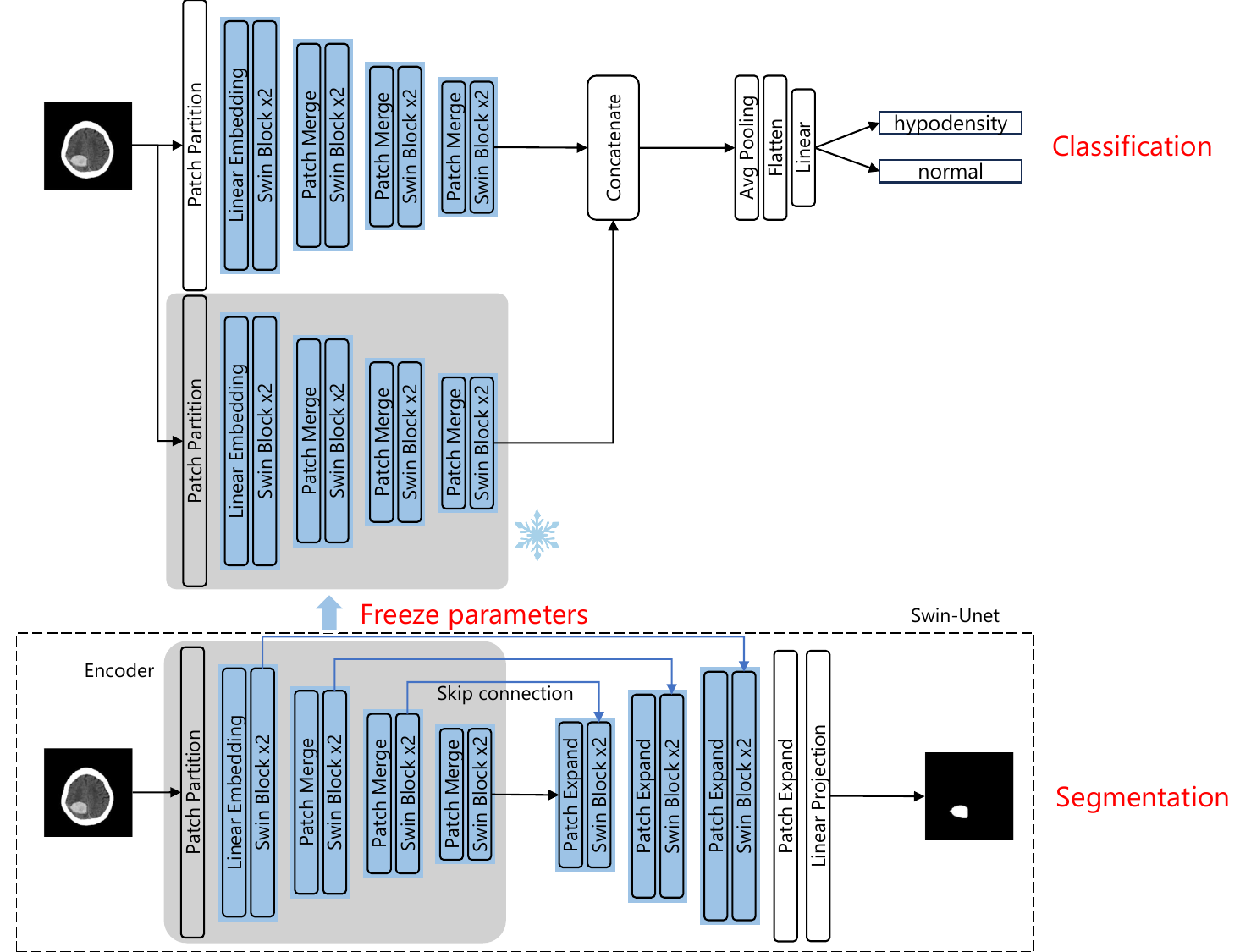}   
\caption{Architecture of Joint-SwinTransformer}
\label{fig:joint-architecture}
\end{center}
\end{figure*}

\paragraph{Encoder with Representation for Segmentation Task}\label{sec:joint-proceedure-1}
As a preliminary segmentation task for joint learning, we trained using Swin-Unet. The encoder of Swin-Unet has the same structure as the SwinTransformer encoder. Within the SwinTransformer blocks, representation learning occurs. The encoder trained by the segmentation task possesses task-specific representations related to segmentation. These representations are then utilized for the classification task.

\paragraph{Encoder during Class Classification}
When performing classification tasks, we utilize both the SwinTransformer encoder and the SwinTransformer encoder trained by the segmentation task. Specifically, the input image is fed into two encoders: one that remains trainable and another that has been frozen after training in Section \ref{sec:joint-proceedure-1}. The feature vectors outputted by each encoder are concatenated along the channel dimension to form a single vector. The patch merging layer in the SwinTransformer encoder serves the role of downsampling. Since patch merging occurs before each stage, if the stage depths are the same, the resolution remains consistent. As both the encoder trained by the segmentation task and the trainable encoder have the same stage depth, the vectors produced by these two encoders are directly concatenated along the channel direction.

\paragraph{Classification Head}
The classification head utilizes the same structure as described in Section \ref{sec:classification head}. However, the input dimension of the linear layer has been modified to accommodate concatenated representations.

\subsubsection{Loss Function}
Similar to the first method, for the preliminary segmentation task, a loss function combining cross-entropy and Dice loss is employed. For the classification task, the cross-entropy function is used.

\section{Experiment}
In this chapter, we conducted experiments using two methods to determine whether the hematoma is hypodense.

\subsection{Dataset}
\subsubsection{Preparing the Dataset}\label{sec:dataset}
In this study, we utilize CT images, which were collected from 11 institutions. Annotation of the extent, location, and condition of hematomas on the CT scans was performed independently by specialists at each of the 11 institutions. The data is stored in DICOM format, and the pixel values are measured in Hounsfield Units (HU). The collected data is contrast-adjusted based on the HU range. Additionally, annotation data for the segmentation task exists for facilities 1 to 4. This data includes masks for pixels where hemorrhages are present.
In acute intracerebral hemorrhage (ICH), markers with different names describe various appearances of hemorrhages, including hypodensity, irregular margins, blend signs, and fluid levels (refer to Figure \ref{fig:hematoma}). These markers, identified by different physicians as potentially useful for ICH, are discussed by Boulouis et al \shortcite{boulouis2016association}. Consequently, the problem involves four overlapping classes, making it a multi-label task. In our experiments, we focus solely on hypodensity and frame the task as classifying whether an image exhibits hypodensity. Specifically, our dataset contains images with no hemorrhage and images that simultaneously belong to all four hemorrhage classes, but we set the task to classify only hypodensity-positive images.

\begin{figure*}[t]
\begin{center}
\includegraphics[width=16cm]{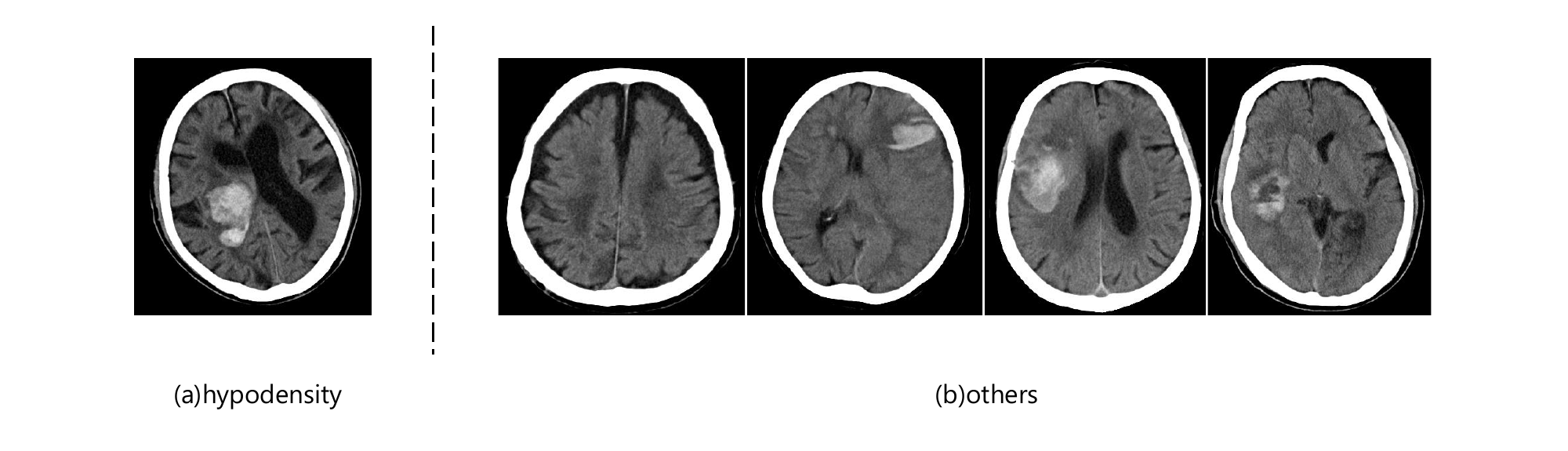}   
\caption{(a) Hypodensity (b) Images without hypodensity overlapping with other hemorrhage classes (from left to right: image without hemorrhage, margin irregular sign, blend sign, fluid level)}
\label{fig:hematoma}
\end{center}
\end{figure*}

\subsubsection{Settings of Two Datasets}
In our experiments, we used a dataset consisting of 11 facilities, totaling 11,780 images. For facility 1, we allocated 179 images as test data, while the remaining 9/10 of facility 1 and 8/10 of the 9,912 images from facilities 2 to 4 were used as training data. The remaining 2/10 of facility 1 served as validation data. Since this dataset involves multiple slices from the same patient’s CT images, there is a possibility of overlap between training and test data due to patient-specific slices. To address this, we randomly selected 179 images from facilities 5 to 11 as additional test data, ensuring that the positive ratio matched that of the test data from facility 1. Specifically, we evaluated two types of test data: ‘Test Data (Hospitals 1-4)’ from facility 1 and ‘Test Data (Hospitals 5-11)’ from facilities 5 to 11. The training data consisted of CT images from patients in hospitals 1-4, while the latter test data did not share slices from the same patients or imaging dates. Consequently, the former scenario does not involve covariance shift when viewed as a classification task, whereas the latter can be considered as having undergone covariance shift.

\subsection{Settings of the Learner} \label{sec:common settings}
In the first proposed method, MTL-Swin-Unet, experiments were conducted under four different settings. Since the classification task is the main task, experiments were performed with three combinations: classification task with image reconstruction task (cls + rec), classification task with segmentation task (cls + seg), and classification task with both segmentation and image reconstruction tasks (cls + seg + rec). Additionally, for the model performing all three tasks (cls + seg + rec), experiments were conducted with an increased model size, referred to as ‘tiny’. In the second proposed method, Joint-SwinTransformer, experiments were conducted with both the normal and tiny sizes. For comparison in the classification task, experiments were conducted using ResNet152 
\cite{DBLP:journals/corr/HeZRS15}, Swin Transformer \cite{liu2021Swin}, Swin Transformer tiny \cite{liu2021Swin}, and CNN based joint-Learning [avoid reference for anonymity]. Hereafter, we call it joint-CNN. For the segmentation task, comparative experiments were conducted using Unet\cite{ronneberger2015u} and Swin-Unet \cite{swinunet}.

For all training, we used Python 3.7 and PyTorch 1.13.1. The training settings mostly followed the guidelines from Swin-Unet. The input image size was resized to \(224\times224\). The encoders for each model were initialized with pre-trained parameters from ImageNet. We employed stochastic gradient descent (SGD) with a momentum of 0.9 and L2 regularization of 0.0001 for training, including 600 epochs for the classification task and 200 epochs for the segmentation task. The initial learning rate and batch size varied across the different tasks. For MTL-Swin-Unet-tiny (cls + seg + rec), we used a batch size of 32 and a learning rate of 0.01. Other multi-task learning models using Swin-Unet and Swin Transformer had a batch size of 64 and a learning rate of 0.01. Joint-SwinTransformer and joint-CNN were trained with a batch size of 128 and a learning rate of 0.004. ResNet was trained with a batch size of 128 and a learning rate of 0.01. Finally, Unet and Swin-Unet were trained with a batch size of 32 and a learning rate of 0.01. We applied learning rate decay to all training runs, where the learning rate \(lr\) decreased according to the following formula at each iteration \(iter\):

\begin{equation}
lr = lr_{base}\times(1-\frac{iter}{iter_{max}})^{0.9}
\end{equation}
Here, \(lr_{base}\) represents the initial learning rate, and \(iter_{max}\) denotes the maximum number of iterations.

Similar data augmentation techniques are applied to both input images and segmentation mask images. In this experiment, data augmentation is achieved by combining the following operations.
\begin{itemize}
      \item After applying random rotations of 0, 90, 180, and 270 degrees, random flips (either vertically or horizontally) are performed
      \item  random rotations within the range of -20 to 20 degrees are applied.
\end{itemize}
Each of these augmentations is independently applied with a 50\% probability. In other words, there are cases where no transformation is applied to the data, cases where only one of the data augmentations is applied, and cases where both augmentations are applied.

Finally, unless explicitly specified, the model’s encoder architecture based on Swin Transformer is configured with 2 blocks in the third stage and an input channel size of 96. The model size of Swin Transformer is determined by these parameters: the number of blocks in the third stage and the input channel count. Specifically, the ‘tiny’ variant has 6 blocks and 96 channels, while the ‘base’ variant has 18 blocks and 128 channels. In the Swin-Unet study, the impact of model size on performance was investigated, but increasing the model size did not lead to significant performance improvements. Therefore, the default setting in the Swin-Unet code uses 2 blocks for the encoder’s third stage, following this configuration when not explicitly specified.

\begin{table*}[h]
\begin{center}
\scalebox{0.9}{
\begin{tabular}{|c|c|ccccc|ccccc|}\hline \hline
           & \multicolumn{6}{|c|}{TestData(Hospital1-4)} & \multicolumn{5}{|c|}{TestData(Hospital5-11)} \\ \hline
   Methods & iou-seg & acc & prec & rec & f1 & auc & acc & prec & rec & f1 & auc \\ \hline
   Unet \cite{ronneberger2015u} & .692 & - & - & - & - & - & - & - & - & - & -  \\ 
   Swin-Unet \cite{swinunet} & .808 & - & - & - & - & - & - & - & - & - & -  \\

   ResNet152 \cite{DBLP:journals/corr/HeZRS15} & - & .901 & .655 & .333 & .692 & .804 & {\bf.822} & .126 & .095 & .504 & .511  \\ 
   Swin Transformer \cite{liu2021Swin} & - & .956 & .813 & {\bf.819} & .895 & .963 & .746 & .254 & .600 & .599 & .772  \\
   SwinTransformer-tiny \cite{liu2021Swin} & - & .952 & .825 & .752 & .880 & .959 & .755 & .240 & .505 & .587 & .741 \\

   joint-CNN (avoid reference for anonymity) & - & .945 & .792 & .724 & .863 & .913 & .787 & .208 & .305 & .560 & .631  \\ \hline
   Joint-SwinTransformer & - & .956 & {\bf.859} & .752 & .888 & {\bf.974} & .797 & {\bf.317} & .638 & {\bf.650} & .787 \\
   Joint-SwinTransformer-tiny & - & .953 & .857 & .724 & .879 & .950 & .782 & .287 & .581 & .626 & .788  \\
   
   MTL-Swin-Unet(cls+rec) & - & .949 & .805 & .743 & .871 & .955 & .736 & .238 & .571 & .586 & .755  \\
   MTL-Swin-Unet(cls+seg) & {\bf.815} & .956 & .823 & .800 & .893 & .967 & .757 & .256 & .648 & .617 & .788  \\
   MTL-Swin-Unet(cls+seg+rec) & .811 & {\bf.961} & .852 & .809 & {\bf.903} & .967 & .756 & .276 & .667 & .618 & {\bf.799} \\
   MTL-Swin-Unet-tiny(cls+seg+rec) & .773 & .960 & .841 & .810 & .901 & .973 & .752 & .282 & {\bf.705} & .623 & .790  \\ \hline
\end{tabular}
}
  \caption{The upper half of the figure displays six models as baseline systems, while the lower half shows six models of the proposed method. The proposed method refers to four models starting with MTL-Swin-Unet, which is a multi-task learning approach, and two models starting with Joint-SwinTransformer. The latter refers to models that transform the convolutional neural network (EfficientNet) used in joint-CNN into a form using SwinTransformer. The left column of test data (Hospitals 1-4) represents cases where non-overlapping slices from the same patient are allowed for both training and test data. Test data (Hospitals 5-11) refers to cases where slices from different patients are used for training and test data. In multi-task learning, there are instances of three-class and two-class multi-tasking, with task names indicating cls: classification task, seg: segmentation task, and rec: image reconstruction task.}
  \label{table:architecture2}
\end{center}
\end{table*}

\subsection{Result}
Table \ref{table:architecture2} presents the results of MTL-Swin-Unet and its comparative methods on the test data from hospitals 1-4 and hospitals 5-11. From the results, MTL-Swin-Unet (cls + seg + rec) achieved the highest accuracy of 0.961 and an F1-score of 0.903 on the test data from hospitals 1-4. For the test data from hospitals 5-11, MTL-Swin-Unet (cls + seg + rec) had a lower F1-score compared to Joint-SwinTransformer, but it achieved the maximum AUC of 0.799.

\paragraph{The Impact of Representations Learned Through Different Tasks on Classification Tasks}
From Table \ref{table:architecture2}, it is observed that MTL-Swin-Unet (cls + seg) surpassed Swin Transformer in AUC on test data (hospitals 1-4) by 0.004 and on test data (hospitals 5-11) by 0.016. Similarly, Joint-SwinTransformer also exceeded Swin Transformer in AUC on test data (hospitals 1-4) by 0.011 and on test data (hospitals 5-11) by 0.015. Conversely, MTL-Swin-Unet (cls + rec) fell behind Swin Transformer in AUC on test data (hospitals 1-4) by 0.008 and on test data (hospitals 5-11) by 0.017. This suggests that the image reconstruction task alone does not share advantageous factors with the classification task and may even contain conflicting information. On the other hand, the segmentation task appears to learn representations beneficial for the classification task. Furthermore, when performing all three tasks of MTL-Swin-Unet (cls + seg + rec), there was no change in AUC on test data (hospitals 1-4) compared to MTL-Swin-Unet (cls + rec), but an improvement of 0.011 on test data (hospitals 5-11) was noted. This indicates that while the image reconstruction task alone does not positively influence the classification task, training it concurrently with the segmentation task enables the learning of representations more advantageous for the classification task than MTL-Swin-Unet (cls + rec).

\paragraph{Impact of Model Size}
Experiments were conducted using MTL-Swin-Unet, Joint-SwinTransformer, and Swin Transformer with the larger ‘tiny’ model size. From Table \ref{table:architecture}, MTL-Swin-Unet-tiny (cls + seg + rec) exhibited a 0.006 increase in AUC compared to MTL-Swin-Unet (cls + seg + rec) on test data from hospitals 1-4. Similarly, Joint-SwinTransformer-tiny showed a 0.001 increase in AUC compared to Joint-SwinTransformer on test data from hospitals 5-11. However, for other cases, a decrease ranging from 0.004 to 0.031 was observed.

\begin{figure*}[h]
\begin{center}
\includegraphics[width=12cm]{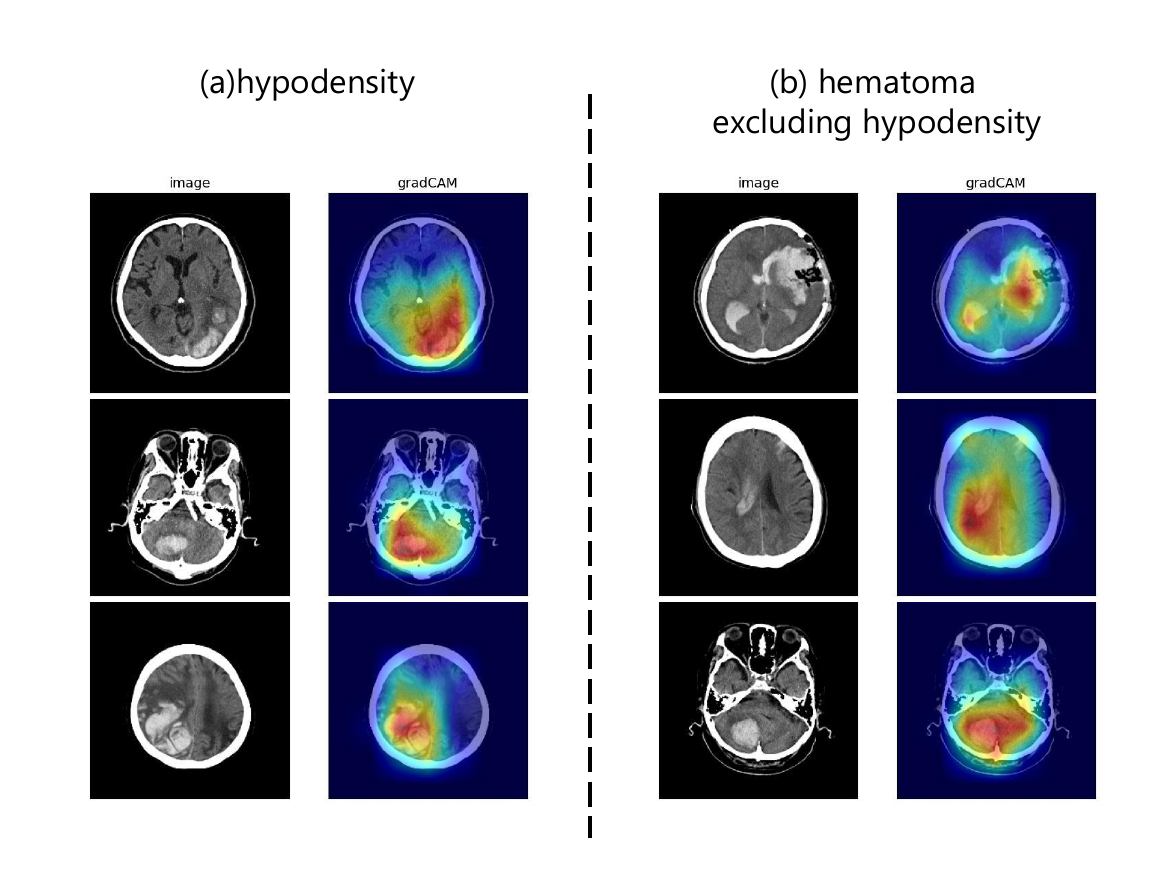}   
\caption{The visualization results of Grad-CAM using MTL-Swin-Unet (cls + seg + rec) for (a) hypodensity images and (b) images with hemorrhage but not hypodensity }
\label{fig:Grad-CAM}
\end{center}
\end{figure*}

\subsubsection{Verification of Prediction Basis}
In brain CT images, models sometimes focus on the skull when making predictions. However, in reality, important information lies in features such as hemorrhages and brain textures. To address this, we employed Grad-CAM (Gradient-weighted Class Activation Mapping) \cite{DBLP:journals/corr/SelvarajuDVCPB16} for visualizing prediction rationales. Grad-CAM generates heatmaps that highlight which regions of an input image most significantly influenced the model’s predictions, providing a visual explanation for the predictions.

Figure \ref{fig:Grad-CAM} presents the visualization results obtained using Grad-CAM for MTL-Swin-Unet (cls + seg + rec). Panel (a) displays the visualization for images with hypodensity, while panel (b) shows the results for images with hematoma but not hypodensity. In both cases, it is evident that the predictions are focused on the hematoma and its surrounding areas.

\section{Conclusion}
In this paper, we propose MTL-Swin-Unet (cls + seg + rec), a multi-task learning approach based on Transformers. The idea behind this method is to address the issue of spurious correlations by leveraging the Transformer’s ability to assist in both segmentation and image reconstruction tasks. By doing so, we aim to obtain more refined representations compared to using only the SwinTransformer classifier. In our experiments, MTL-Swin-Unet (cls + seg + rec) outperformed other classifiers in terms of F1-score when the test data included slices from the same patients (a non-covariate shift setting). Conversely, in the setting where the test data did not include slices from the same patients (a covariate shift setting), MTL-Swin-Unet (cls + seg + rec) achieved the highest AUC.

\begin{table*}[h]
\begin{center}
\scalebox{0.9}{
\begin{tabular}{|c|ccccc|}\hline
   Methods & acc & prec & rec & f1 & auc \\ \hline
   
   ResNet & .901 & .655 & .333 & .692 & .804  \\ 
   Swin Transformer & .956 & .813 & {\bf.819} & .895 & .963  \\

   joint-CNN & .945 & .792 & .724 & .863 & .913 \\ \hline
   Ours(Joint-SwinTransformer) & .956 & {\bf.859} & .752 & .888 & {\bf.974} \\
   
   Ours(cls+rec) & .949 & .805 & .743 & .871 & .955  \\
   Ours(cls+seg) & .956 & .823 & .800 & .893 & .967  \\
   Ours(cls+seg+rec) & {\bf.961} & .852 & .809 & {\bf.903} & .967 \\ \hline
\end{tabular}
}
  \label{table:architecture}
\end{center}
\end{table*}


\bibliography{aaai25}

\end{document}